\documentclass[10pt,twocolumn,letterpaper]{article}

\usepackage{wacv}
\usepackage{times}
\usepackage{epsfig}
\usepackage{graphicx}
\usepackage{amsmath}
\usepackage{amssymb}

\usepackage{algorithm}
\usepackage{algorithmic}
\usepackage[ruled,vlined,algo2e]{algorithm2e}
\usepackage{helvet} 
\usepackage{courier}  
\usepackage[numbers]{natbib}
\usepackage[utf8]{inputenc} 
\usepackage[T1]{fontenc}    

\usepackage{booktabs}       
\usepackage{amsfonts}       
\usepackage{nicefrac}       
\usepackage{microtype}      

\usepackage{epsfig}
\usepackage{adjustbox}

\usepackage{amsthm}  
\usepackage{wasysym}

\usepackage{color}
\usepackage{dsfont}	
\usepackage{wrapfig}
\usepackage{footnote}
\usepackage{epstopdf}
 \usepackage{enumitem}
\usepackage{subfigure}
\usepackage{caption}
\usepackage{comment}
\usepackage{multirow}
\usepackage{authblk}

\def\eg{\emph{e.g., }}
\def\ie{\emph{i.e., }}

\def\etal{\emph{et al. }}

\usepackage{pifont}

\usepackage{mathtools}

\makeatletter
\newcommand*{\rom}[1]{\expandafter\@slowromancap\romannumeral #1@}
\makeatother
\makeatletter
\newcommand\footnoteref[1]{\protected@xdef\@thefnmark{\ref{#1}}\@footnotemark}
\makeatother
\newcommand{\bfsection}[1]{\vspace*{0.1cm}\noindent\textbf{#1.}}

\wacvfinalcopy 

\ifwacvfinal
\usepackage[breaklinks=true,bookmarks=false]{hyperref}
\else
\usepackage[pagebackref=true,breaklinks=true,colorlinks,bookmarks=false]{hyperref}
\fi

\begin{document}

\title{EllipsoidNet: Ellipsoid Representation for Point Cloud Classification and Segmentation}

\author[1,2]{Yecheng Lyu}
\author[1]{Xinming Huang}
\author[1]{Ziming Zhang}
\affil[1]{Worcester Polytechnic Institute}
\affil[2]{Volvo Car Technology USA}
\affil[ ]{\tt\small yecheng.lyu@volvocars.com, \{xhuang,zzhang15\}@wpi.edu}

\maketitle

\begin{abstract}
Point cloud patterns are hard to learn because of the implicit local geometry features among the orderless points. In recent years, point cloud representation in 2D space has attracted increasing research interest since it exposes the local geometry features in a 2D space. By projecting those points to a 2D feature map, the relationship between points is inherited in the context between pixels, which are further extracted by a 2D convolutional neural network. However, existing 2D representing methods are either accuracy limited or time-consuming. In this paper, we propose a novel 2D representation method that projects a point cloud onto an ellipsoid surface space, where local patterns are well exposed in ellipsoid-level and point-level. Additionally, a novel convolutional neural network named EllipsoidNet is proposed to utilize those features for point cloud classification and segmentation applications. The proposed methods are evaluated in ModelNet40 and ShapeNet benchmarks, where the advantages are clearly shown over existing 2D representation methods. 
\end{abstract}

\section{Introduction}
    Point cloud is a widely used data structure that represents the 3D objects using an orderless list of scanned points. In recent years, tremendous research efforts have been conducted on point cloud pattern recognition in the field of automated driving \cite{chen2019suma++,lyu2018real,wu2018squeezeseg,lang2019pointpillars,yang2018pixor,milioto2019rangenet++,lei2019octree} and indoor/outdoor scene recognition \cite{qi2017pointnet,qi2017pointnet++,thomas2019kpconv}. As concluded in Guo \etal{}\cite{guo2020deep}, 2D representation learning is one of the promising approaches to extract local and global features from point clouds. 
    2D representations, including multi-view representation \cite{su15mvcnn}, spherical representation \cite{milioto2019rangenet++,chen2019suma++,zheng2020lodonet,lyu2018real,wu2018squeezeseg}, and bird-eye-view representation \cite{yang2018pixor,lang2019pointpillars}, are applied to point cloud learning tasks because they benefit from 2D convolutional neural networks (CNNs), which are well studied in image learning tasks \cite{he2016deep,ronneberger2015unet,kirillov2019panoptic,lin2017fpn}. However, existing 2D representations methods either require point clouds from specific sensors (\eg{} LiDARs \cite{lyu2018real,milioto2019rangenet++,wu2018squeezeseg} ) or have huge point loss rate without a sufficient study of impact to related applications \cite{su15mvcnn,Meng_2019_VV_Net}. Recent methods try to overcome those drawbacks by mapping the points near-losslessly \cite{lyu2020learning} or mapping only local regions around target point in each time \cite{lin2020fpconv}, which however are time inefficient. Thus, it is still an unsolved problem to apply an effective and efficient representation to a point cloud.
    
    \begin{figure}[t]
    \centering
    \includegraphics[width=0.9\columnwidth]{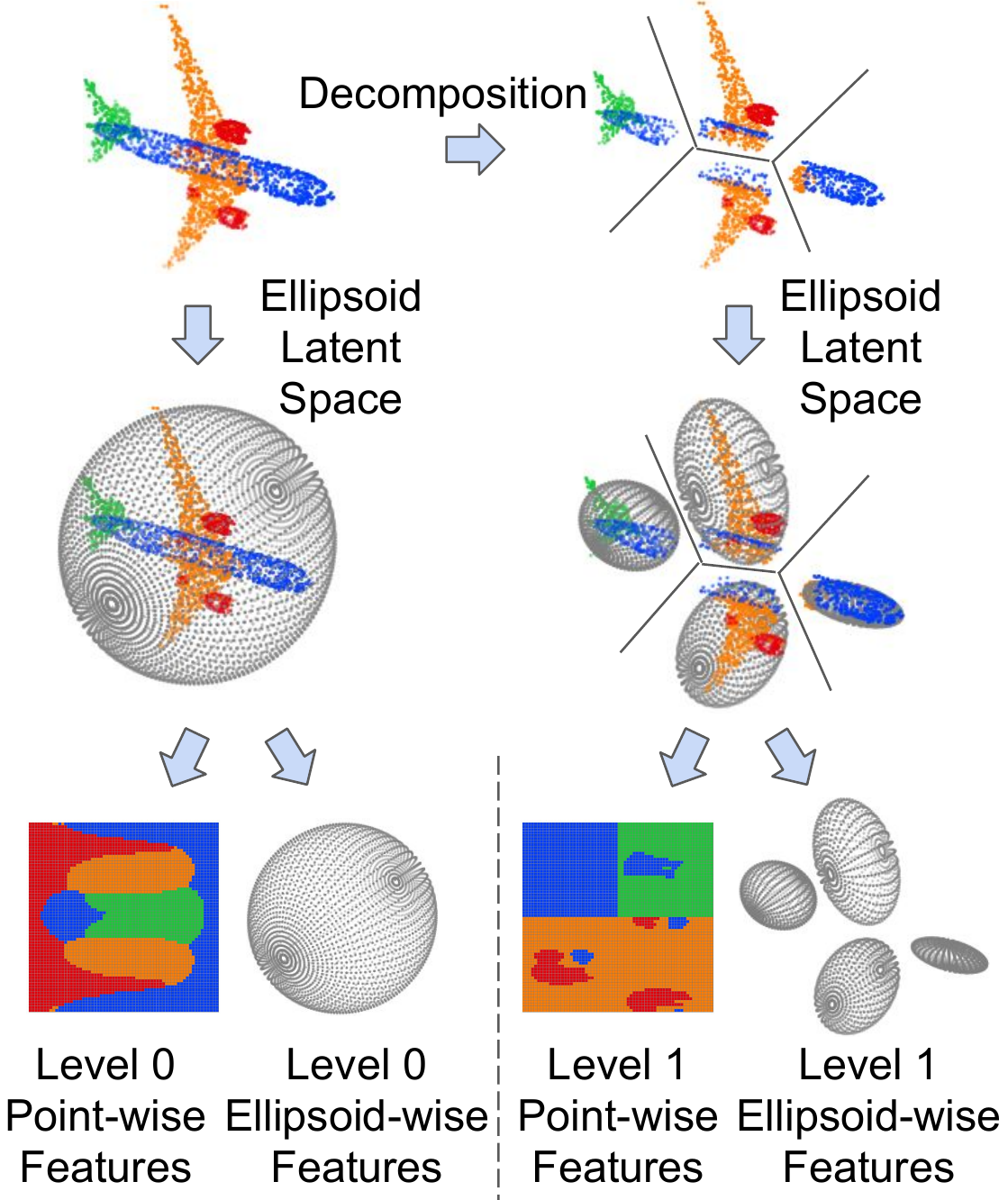}
    \caption{Overview of EllipsoidNet. \label{fig:ellpisoidnet_intro}}
    \vspace{-0.5cm}
    \end{figure}
    
\bfsection{Motivation}
    Our work is motivated by the spherical representation, a popular representation method for LiDAR point cloud processing \cite{lyu2018real,lyu2018chipnet,milioto2019rangenet++,zheng2020lodonet,wu2018squeezeseg,chen2019suma++}. In LiDAR point cloud frames, points are near-uniformly distributed on a spherical space so that a LiDAR point cloud is projected to a 2D feature map with minimal point overlaps and void pixels in the feature map. Lyu \etal{} \cite{lyu2018chipnet} compares several representation method on KITTI road benchmark \cite{Fritsch2013KITTI_road} and finds that the spherical representation of a LiDAR frame is a well balanced solution between point usage and latent space usage. In general point cloud processing, however, a spherical representation may not work well since the points are usually not uniformly distributed in the spherical space. 
    
    In our work, we extend the use of spherical representation in general point cloud feature extraction. The proposed representation method effectively organizes the points in a 2D latent space with minor point overlaps and minor limitations to the network processing. Additionally, the proposed representation method is time efficient comparing to existing methods \cite{su15mvcnn,wang2019dynamic,li2018pointcnn,liu2019rscnn}.

\bfsection{Approach}
    To address the spherical representation of general point cloud issue, we extend the spherical representation to hierarchical ellipsoid representation by introducing the following schemes: (1) point cloud decomposition, (2) spatial transform, and (3) feature map completion. 

    Point cloud decomposition aims to split a big point cloud to several point partitions, in which local patterns can be better recognized and time efficiency can be improved. Spatial transform aims to generate a ellipsoid latent space that better fits the point partitions. Feature map completion is used to fill each pixel in the latent space with reasonable values to generate a dense feature map for further processing.
    
    The ellipsoid representation of a point cloud outputs two items: a combination of 2D feature maps for point-wise features and a combination of feature vectors for ellipsoid-wise features. Targeting the classification and semantic segmentation applications, we further propose EllipsoidNet, a neural network based solution to process the ellipsoid representation of a point cloud. The network utilizes both the point-wise features and ellipsoid-wise features for point cloud classification and segmentation.

\bfsection{Performance preview}
    To demonstrate the effectiveness of our ellipsoid representation, we evaluate the proposed method in two point cloud datasets: ModelNet40 for classification, and ShapNet for segmentation. In the experiment, the proposed method achieves near state-of-the-art performance in both datasets, with the help of the proposed neural networks.

\bfsection{Contributions}
    In summary, our key contributions of this paper are as follows:
    \setlist[itemize]{leftmargin=*}
    \begin{itemize}[nosep]
    \item A novel representation method is proposed that represent point cloud in ellipsoid latent space. This method project a point cloud in dense 2D feature maps and works hierarchically for large-scale point cloud.   Additionally, we explicitly present its impact to point cloud classification and segmentation application.
    
    \item Accordingly, a novel neural network named EllipsoidNet is proposed to take advantage of the ellipsoid representation for point cloud classification and segmentation applications.
    
    \item We evaluate the proposed representation method and networks in a point cloud classification dataset and a point cloud segmentation datasets, and present its effectiveness and efficiency.
    \end{itemize}
    
\section{Related work}
\bfsection{Point cloud decomposition} 
    Point cloud decomposition is widely used to separate the points to small partitions, reduce the computational complexity of point cloud learning, and enhance local pattern recognition. Lyu \etal{} \cite{lyu2020learning} utilizes a modified Kmeans clustering to create balanced point cloud partitions, which significantly decrease the computational complexity of point cloud representation. ACD \cite{lien2007acd} proposed an approximate convex decomposition method that divides an 3D object into near-convex parts, which however requires object surface meshes and unable to apply directly to point clouds. Cvxnet \cite{Deng2020cvxnet} introduces a network architecture to represent a low dimensional family of convexes. Yu \etal{}\cite{yu2019partnet} performs recursive binary decomposition to a point cloud and concatenate the features from all tree levels. $\Psi$-CNN \cite{lei2019octree}, Kd-net \cite{klokov2017escape}, and TreeRNN \cite{Lyu2020TreeRNN} perform similar decomposition schemes for point cloud and graph learning tasks. In this paper, we take the Kmeans for point cloud decomposition since it is fast and generates a fixed number of point partitions.

\bfsection{Spatial transform}
    Spatial transform is used to normalize a point cloud and learn a robust feature extractor. Spatial transform is usually achieved by applying a 3D transform to each point, which results in 3D rotation, translation, or scaling to the target point cloud. PointNet \cite{qi2017pointnet} and PointNet++ \cite{qi2017pointnet++} apply two spatial transforms to the point cloud: a non-learnable transform that scale and transit the point cloud within a uniball, and a learnable transformer named T-net for better feature extraction. In the experiment, they show that T-net has a significant improvement to point cloud classification. Wang \etal{} \cite{wang2019spatial} proposes a novel approach to learn different non-rigid transformations of a input point cloud for different local neighborhoods at each layer. They claim that spatial transformers can learn the features more efficiently by altering local neighborhoods according to the semantic information of 3D shapes regardless of variations in a category.

\bfsection{Feature map completion}
    Feature map completion is used to generate a dense feature map with no void pixels, which helps CNNs better extract local features. This method is frequently used for depth completion in LiDAR point cloud processing. FuseNet \cite{chen2019learning} learns to extract joint 2D and 3D features for depth completion, which results in a dense depth map from camera and LiDAR inputs. LodoNet \cite{zheng2020lodonet} completes the depth map using its nearest valid depth values. 
    Different from the nearest neighbourhood filling, InterpCNN \cite{mao2019InterpCNN} and TangentConv \cite{tatarchenko2018TangentConv} perform the completion using Gaussian mixture.
    In other point cloud learning applications, FoldingNet \cite{yang2018foldingnet} and AtlasNet \cite{groueix2018atlasnet} try to learn a latent vector space that cover the point cloud with 2D surfaces.
    In this paper, we propose to fill a dense feature map by projecting the nearest point in the point cloud to anchors in the ellipsoid latent space. 

\bfsection{Point cloud 2D representation learning}
    Point cloud 2D representation learning tries to project a point set onto a 2D latent space, from which 2D CNNs are adopted for feature extraction. Besides the methods that focus on LiDAR point cloud only \cite{lyu2018real,milioto2019rangenet++,wu2018squeezeseg}, several universal methods are proposed to represent a point cloud in 2D space. There are two major approaches: projection based and learning based. Projection based approaches try to define a 2D surface in the 3D space and then projects the points to the surface. MVCNN \cite{su15mvcnn} projects the points into multiple 2D plane space and concatenates the features from all 2D plane representations for point cloud classification. Similarly, in DeePr3SS \cite{lawin2017DeePr3SS}, SnapNet \cite{boulch2018snapnet} and TangentConv \cite{tatarchenko2018TangentConv}, the input point cloud is projected into multiple virtual camera views. Lyu \etal{} explores the use of graph drawing to project the point cloud into a single 2D space for segmentation purposes. To better represent the local geometry pattern in 2D space, FPConv \cite{lin2020fpconv} proposes a learnable method that projects the local region of each point onto a 2D image space in each layer.  
    
\bfsection{Other point cloud learning methods}
    Besides the 2D representation learning, volumetric, graph, and structure are also widely used methods for point cloud representation learning. Volumetric-based methods \cite{Meng_2019_VV_Net,zhou2018voxelnet,wang2017ocnn,graham2018SparseConvNet} discretize the 3D space to voxels, to which 3D convolution based methods can be applied. Graph-based methods \cite{wang2019dynamic,simonovsky2017ecc,landrieu2018spg} connect the points to graphs and apply graph convolutions and poolings to them. Structure-based methods \cite{yu2019partnet,klokov2017kdnet,li2018sonet} decomposite a point cloud layer by layer into a tree, and apply point-wise convolutions and recurrent neural networks to it.

\section{Ellipsoid representation of point cloud} \label{sec:EllipsoidNet}

\subsection{Method overview}
    Given a point cloud $\mathcal{P} \in \mathcal{R}^{N\times3}$ and a 2D feature map $\mathcal{I} \in \mathcal{R}^{M \times M \times C}$ where $N$, $M$, $C$ denote the point size, the feature map resolution and feature size, the ellipsoid representation of point cloud is to perform a projection $f: \mathcal{P} \rightarrow \mathcal{I}$ that maps the point cloud to a 2D latent space. Figure \ref{fig:ellpisoid_representation_intro} illustrates the steps of this representation. For a large-scale point cloud with a complex shape, we further propose a hierarchical pipeline that performs the representation in multiple scales. 
    
    In this section, we firstly introduce the method in detail; secondly, we present the hierarchical scheme of this method for a large-scale point cloud with a complex shape; at last, we analysis the impact of this representation to point cloud classification and segmentation applications and illustrate its effectiveness and efficiency.

\subsection{Approach}
    Given a point cloud $\mathcal{P}$, the spherical representation is equivalent to constructing its circumsphere, mapping each point to it's nearest point on the sphere, and generating a feature map $I$ by sampling on the circumsphere. However, this method has two disadvantages: (1) unbalanced projection, and (2) sparse feature map. Some point clouds are much longer radii along their major axis than that along the minor axis, which results in heavily overlapped projection onto the poles of the circumsphere and many void pixels in the other parts; the pixels in feature map are not always filled with 3D points, which results in void pixels that confuse the CNNs .

\begin{figure}[t]
        \centering
        \includegraphics[width=0.85\columnwidth]{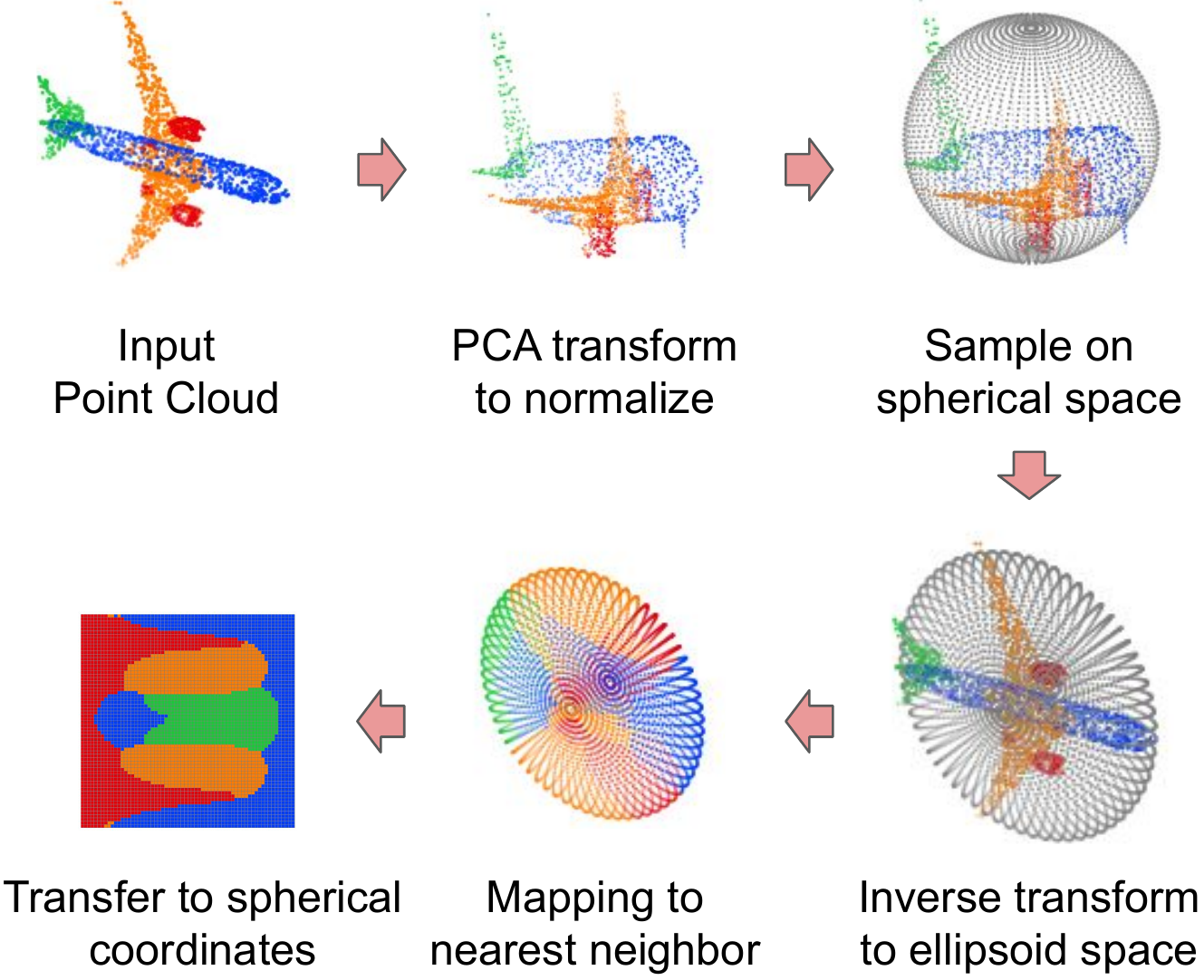}
        \caption{Ellipsoid representation of a point cloud. \label{fig:ellpisoid_representation_intro}}
        \end{figure}

    Tackling this two issues, we proposed a novel representation method that generates a dense feature map and reduce the overlaps. As illustrated in Figure \ref{fig:ellpisoid_representation_intro}, the method has two upgrades: (1) use ellipsoid surface instead of circumsphere, and (2) feature-map-orientated projection. The first upgrade is achieved by measuring the ellipsoid radius of the point cloud using the principal component analysis (PCA), and then scaling the circumsphere by the inverse of the radius. In this way, an ellipsoid surface of the point cloud is constructed that better fits the 3D shape. The second upgrade is achieved by tracing the nearest point of each pixel in the latent feature map and fill them into the pixel. By switching from point-cloud-orientated projection scheme to feature-map-orientated scheme, each point can be projected to multiple feature map pixels and all pixels can be filled with points. The algorithm is described in Algorithm \ref{Alg:Ellipsoid_representation}.
    
    By applying the ellipsoid representation, we successfully build the correspondence between the points in 3D space and the pixels in the latent space.

    \begin{algorithm}[t]
        \KwIn{Point Cloud $\mathcal{P}$, feature map resolution $M$}
        \KwOut{Ellipsoid-wise feature vector $F_e$, Point-wise feature map $F_p$}
        \quad \\
        $\bf{V} \leftarrow PCA(\mathcal{P})$; \\
        $\mathcal{P}_1 \leftarrow \mathcal{P}\bf{V}^T$;\\
        $radii_{xyz} = (\mathcal{P}_{1,xyz}^{max} - \mathcal{P}_{1,xyz}^{min})/2$;\\
        $t \leftarrow (\mathcal{P}_{1,xyz}^{max}/2 + \mathcal{P}_{1,xyz}^{min}/2)\bf{V}$;\\
        $rot \leftarrow RotVec(\bf{V})$;\\
        $F_e \leftarrow [rot,radii,t]$;\\
        \textbf{FOR} $u,v \in 0,1,2,...,M$:\\
        $p_{Sphere} \leftarrow Spherical\_to\_Cartesian(u/M,v/M)$;\\
        $Anchor \leftarrow (p_{Sphere}*radii)\bf{V}-t$;\\
        $p(u,v) \leftarrow NearestNeighbour(Anchor) \in \mathcal{P}$; \\
        $F_p(u,v) \leftarrow [p(u,v),p_1(u,v)\in \mathcal{P}_1,p_{Sphere}(u,v),(u,v)]$;\\
        \textbf{ENDFOR}
        
        \textbf{Return} $F_e,F_p$
        
    	\caption{Ellipsoid representation \label{Alg:Ellipsoid_representation}}
    \end{algorithm}

\bfsection{Point-wise features and ellipsoid-wise features}
    The key novelty of the ellipsoid representation is that it outputs not only point-wise features but also ellipsoid-wise features. The point-wise features are organized as a feature map where extensive local features are included. The ellipsoid-wise features are organized as a feature vector containing the global features of the target point cloud.
    
    Point-wise feature includes the point's $x,y,z$ position in the 3D space, its relative position in the ellipsoid coordinate frame, and the pixel's position in the 3D space and its relative position in the ellipsoid coordinate frame. Since we have the correspondence between the points and the pixels, we manage to organize a 2D feature map $\mathcal{I} \in \mathcal{R}^{M \times M \times C}$, where each pixel is a feature vector of point-wise features. Ellipsoid-wise feature contains the position, orientation and radius of the ellipsoid that covers the point cloud.

\subsection{Hierarchical representation}
    Ellipsoid representation is an effective way to extract local and global features from a point cloud. However, it is still difficult to represent a point cloud with a complex shape. To this end, we further proposed a hierarchical pipeline to generate the ellipsoid representation in multiple scales.
    
    Figure \ref{fig:ellpisoidnet_intro} illustrates the steps of a 2-level hierarchical representation. Given a point cloud $\mathcal{P}$, we decomposite it into $N_c$ partitions denoted as $P_1, P_2, ..., P_{N_c}$. After decomposition, we generate the ellipsoid representation of each partition. In multi-scale representation, we further decomposite each partition to even smaller partitions and generate ellipsoid representation (\ie{} decomposite $P_1$ to $P_{11}, P_{12}, ..., P_{1N_c}$).
    In that way, we can extract point-wise features and ellipsoid-wise features in multi-scales and feed them all into the neural network.
    
    In this paper, we utilize the Kmeans solver for point cloud decomposition because it generates compact partitions and is time efficient.

\subsection{Analysis}
\begin{figure}[t]
    \centering
    \includegraphics[width=0.9\columnwidth]{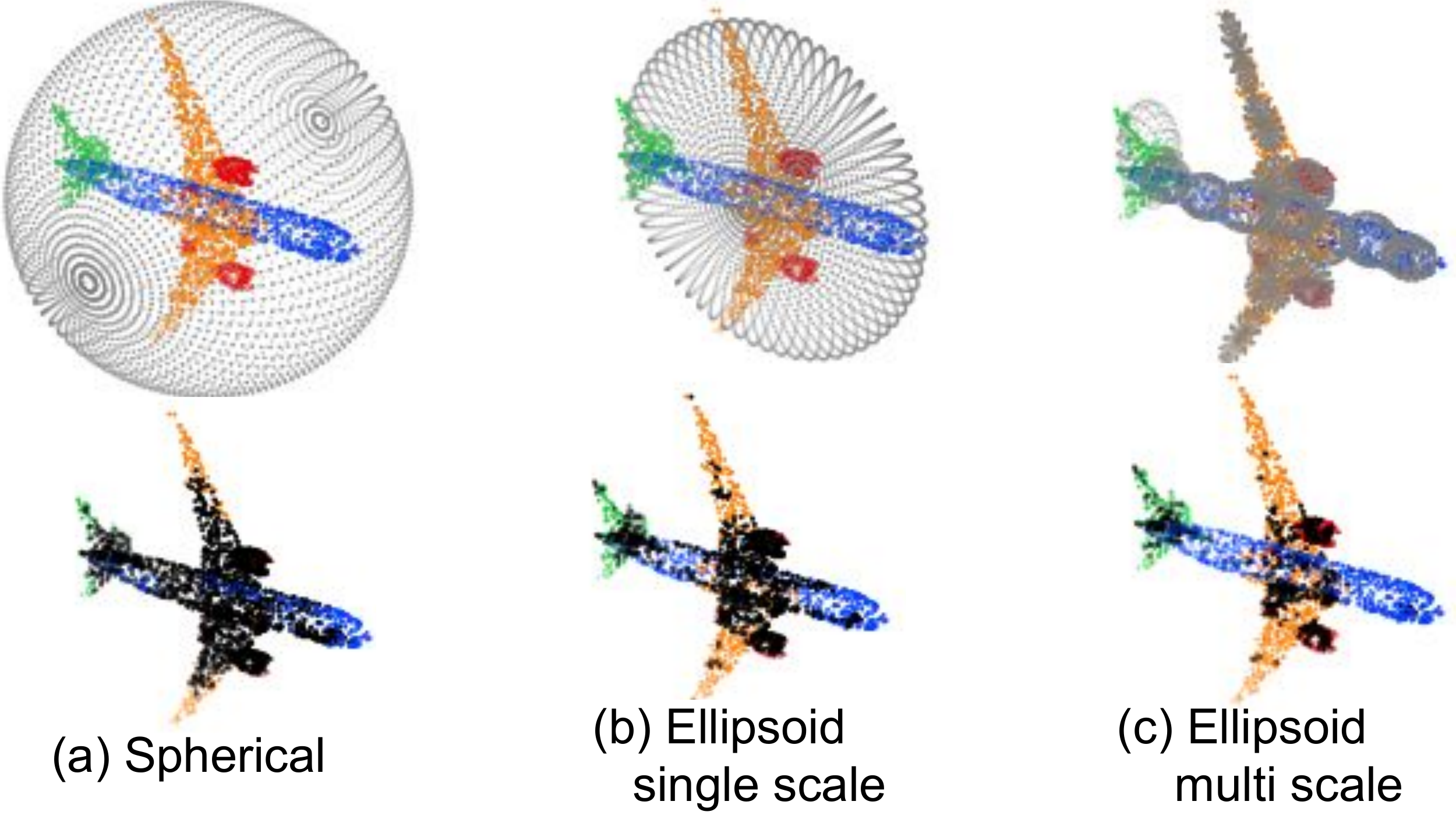}
    \caption{Qualitative evaluation of ellipsoid representation. The upper row shows the constructed latent space, and the lower row shows the projected points (colored by semantic segments) and missed points (black). \label{fig:qualitative}}
    \end{figure}

    To analysis the performance of our ellipsoid representation, we manage to evaluate its impact to classification and segmentation on ShapeNet dataset. For qualitative evaluation, we test on an airplane point cloud and visualize results of spherical, single-scale ellipsoid and multi-scale ellipsoid representation. In Figure \ref{fig:qualitative} our multi-scale ellipsoid representation better fits the point cloud, and has much more points mapped to the latent space.

    \begin{figure*}[t]
    \centering
    \includegraphics[width=0.9\textwidth]{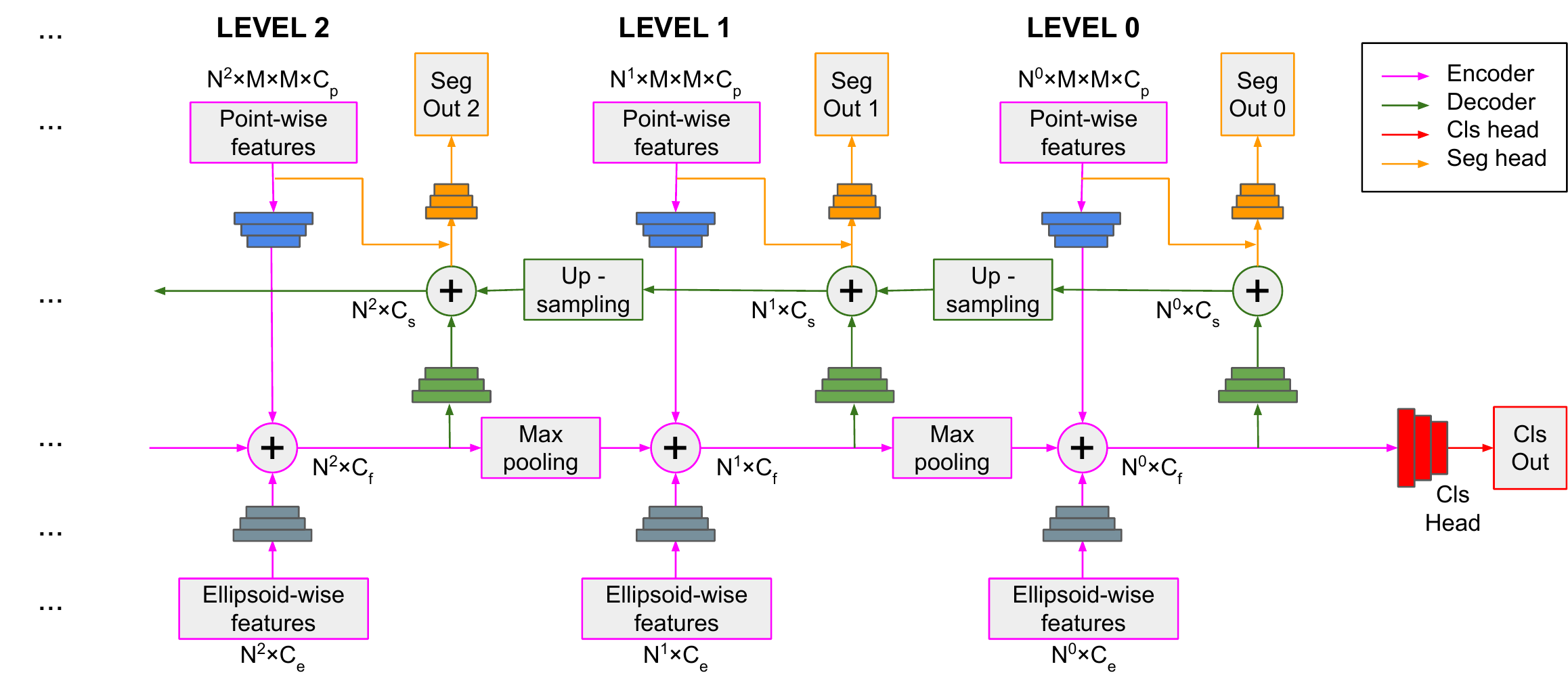}
    \caption{Illustration of our EllipsoidNet architecture. \label{fig:EllipsoidNet_Overview}}
    \end{figure*}
\bfsection{Impact to classification and segmentation}
    To quantitatively evaluate the impact of ellipsoid representation to point cloud classification and segmentation, we generate the representations for all the
    14007 point cloud objects in the ShapeNet training set and compare the point usage rate and maximum segmentation intersect over union (IoU). The point usage rate measures the percentage of points in the point cloud that are mapped to the feature maps. A representation with high point usage ratio is expected to keep more local pattern, which may further leads to better performance in classification and segmentation. The maximum segmentation IoU measures the instance mean IoU of the dataset if we have correct segment on feature maps. This metric values the upper bound of the segmentation performance. 
    
    In Table \ref{tab:point_usage}, we compare the between different representations with the same feature map size. We observe that our multi-scale ellipsoid representation has a significant improvement towards single-scale ellipsoid representation and spherical representation, which indicates that our proposed method potentially performs better in the point cloud applications. Our experiments in Section \ref{sec:Experiment} further confirm this point.
    
    In Table \ref{tab:point_usage_decomposition} we compare between multi-scale representation with different partition sizes, and in Table \ref{tab:point_usage_imreso} we compare between representation with different feature map sizes. Clearly we see that more partitions and larger those metrics. 

    \setlength{\tabcolsep}{2pt}
    \begin{table}[t]
    		\centering
    		\begin{tabular}{c|c|c|c}
    			\toprule
    			Representation & Spherical & \begin{tabular}[c]{@{}c@{}} Ellipsoid \\ single-scale\end{tabular} & \begin{tabular}[c]{@{}c@{}}Ellipsoid\\multi-scale\end{tabular} \\
    			\midrule
    			\begin{tabular}[c]{@{}c@{}} Feature map size\end{tabular}  & $64^2$ & $64^2$ & $16\times16^2$\\
    			\midrule
    			\begin{tabular}[c]{@{}c@{}} Point usage rate (\%)\end{tabular} & 21.6 & 41.1 & 72.9 \\
    			\midrule
    			\begin{tabular}[c]{@{}c@{}c@{}} Max seg. IoU (\%)\end{tabular} & 82.1 & 88.0 & 95.3 \\

    			\bottomrule
    		\end{tabular}
    	\caption{Comparison among different representations.\label{tab:point_usage}}
    	  \vspace{-2mm}
    \end{table}
    
    \setlength{\tabcolsep}{2pt}
    \begin{table}[t]
    		\centering
    		\begin{tabular}{c|c|c|c}
    			\toprule
    			Representation & \multicolumn{3}{c}{Ellipsoid multi-scale}\\
    			\midrule
    			\begin{tabular}[c]{@{}c@{}} Feature map size\end{tabular}  & $16\times16^2$ & $25\times16^2$ & $36\times16^2$\\
    			\midrule
    			\begin{tabular}[c]{@{}c@{}} Point usage rate (\%)\end{tabular} & 72.9 & 78.3 & 86.2 \\
    			\midrule
    			\begin{tabular}[c]{@{}c@{}c@{}} Max seg. IoU (\%)\end{tabular} & 95.3 & 96.2 & 97.4 \\

    			\bottomrule
    		\end{tabular}
    	\caption{Comparison between ellipsoid representation methods with different decomposition sizes.\label{tab:point_usage_decomposition}}
    	  \vspace{-2mm}
    \end{table}

    \begin{table}[t]

    		\centering
    		\begin{tabular}{c|c|c|c}
    			\toprule
    			Representation & \multicolumn{3}{c}{Ellipsoid multi-scale}\\
    			\midrule \begin{tabular}[c]{@{}c@{}} Feature map size\end{tabular}  & $36\times16^2$ & $36\times32^2$ & $36\times64^2$\\
    			\midrule
    			\begin{tabular}[c]{@{}c@{}} Point usage rate (\%)\end{tabular} & 86.2 & 90.1 & 91.3 \\
    			\midrule
    			\begin{tabular}[c]{@{}c@{}c@{}} Max seg. IoU (\%)\end{tabular} & 97.4 & 97.9 & 98.1 \\

    			\bottomrule
    		\end{tabular}
    	\caption{Comparison between ellipsoid representation methods with different feature map sizes.}
    	
    	\label{tab:point_usage_imreso}
    		
    	  \vspace{-2mm}
    \end{table}



\section{EllipsoidNet}\label{sec:Network}


\subsection{Network overview}

    Figure \ref{fig:EllipsoidNet_Overview} overviews of the EllipsoidNet for classification and segmentation applications. The network contains two major flows (encoder and decoder) and two major heads (classification head and segmentation head). The encoder extracts the features from the ellipsoid representations, the decoder distributes the extracted features back to the local regions, and the heads focus on specific tasks. For multi-scale ellipsoid representation there are several levels of point-wise feature maps and ellipsoid-wise feature vectors. To accept those feature inputs, in each level, there is a encoder to parse the point-wise features and another one for ellipsoid features. We further fuse the encoder outputs in each level by adding them together, and then send them to the next level after a max-pooling. In classification tasks, we add a classification head at the end of the encoder, and in segmentation tasks, we add both classification head and segmentation head.
    
    The input of the EllipsoidNet are point-wise features and ellipsoid features in all hierarchical levels. For classification tasks, the network outputs the object class estimation; for segmentation tasks, the network outputs object class estimation as auxiliary loss, and point-wise segmentation estimates in all levels.

\subsection{Modules}
    \begin{figure}[t]
    \centering
    \includegraphics[width=0.9\columnwidth]{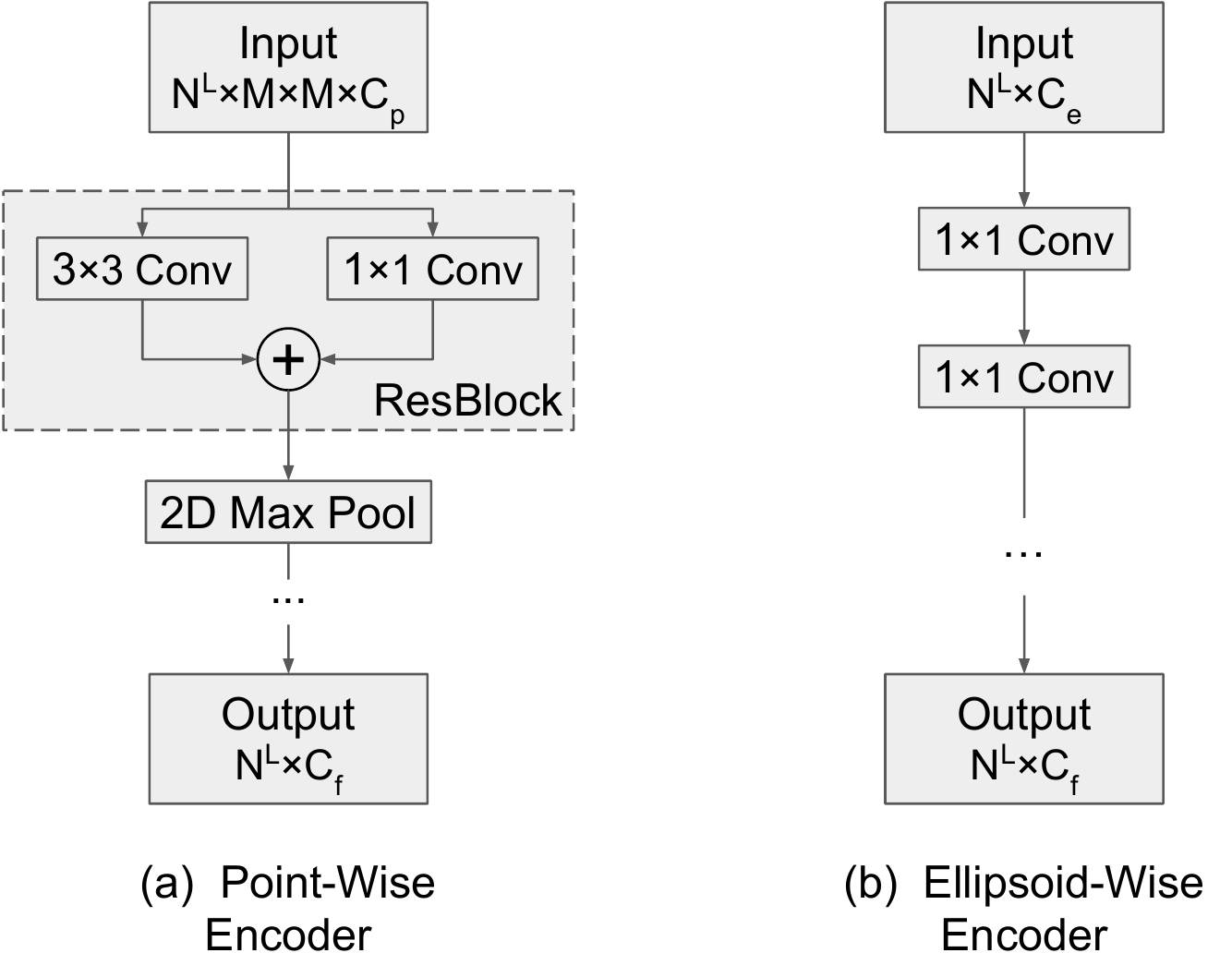}
    \caption{Modules of EllipsoidNet. \label{fig:EllipsoidNet_Modules}}
    \vspace{-0.5cm}
    \end{figure}
    
\bfsection{Point-wise feature encoder}
    This encoder extracts the local features from point-wise feature maps. As presented in Figure \ref{fig:EllipsoidNet_Modules} (a), we employ ResBlock inspired from ResNet \cite{he2016deep} to extract local features, and 2D max-pool for downsampling. By adding multiple ResBlocks and pooling layers, we successfully extract $C_f$ features from each point-wise feature map with size $M\times M\times C_p$. Feature maps from each point partition is processed in parallel.  

\bfsection{Ellipsoid-wise feature encoder}
    This encoder fuses the ellipsoid-wise features in each point partition. As shown in Figure \ref{fig:EllipsoidNet_Modules}(b), we employ $1\times 1$ convolution layers similar to PointNet \cite{qi2017pointnet}. The encoder outputs are directly added to the output of the point-wise feature encoder and the outputs of lower levels. Similar to the point-wise feature encoder, the feature vectors from each partition is processed in parallel.

\bfsection{Classification head}
    The classification head estimates the category of an point cloud object based on the extracted features. In point classification tasks, its output is the final result; in point segmenting tasks, its output serves as an intermediate result that helps the network training. This head is constructed by several fully-connected layers.

\bfsection{Segmentation head}
    The segmentation head estimates the category of each point in a point cloud. This head is not implemented for point cloud classification tasks. As seen in Figure \ref{fig:EllipsoidNet_Overview}, this head returns a segmentation map in each level, where we further estimate the point category. The segmentation head contains several multi-layer perceptrons (MLPs).
    
\bfsection{From network output to point cloud segmentation}
    The segmentation head outputs a segmentation map for each level. In order to get the point cloud segment from the map segment, we trace back the projected points from map segment according to the ellipsoid representation. For the points that is projected to multiple pixels in the segment map, we use the mean value of all the estimation. For minor points that are not projected to the segment map, we assign them according to their nearest neighbour in the point cloud with known segment labels. In this way the semantic labels of all points are estimated.

\section{Experiment}\label{sec:Experiment}

\subsection{Datasets}
    The proposed ellipsoid representation and EllipsoidNet are evaluated in two dataset: ModelNet40 for classification, and ShapeNet for segmentation.

    ModelNet40 dataset contains a 9840-object training set and a 2468-object testing set. Each object belongs to one of 40 classes and is sampled to 2048 points. Similarly, ShapeNet dataset contains a 14007-object training set and a 2874-object testing set, where each object belongs to one of 16 object classes and is sampled to 2048 points. In addition, each point in the dataset belongs to one of the 50 semantic class.
    
    The metrics for ModelNet40 classification benchmark are overall accuracy (OA) and mean class accuracy (mAcc), and the metric for ShapeNet segmentation benchmark are the mean intersection over union (mIoU) in class level (class mIoU) and object level (instance mIoU).
    
\subsection{Ellipsoid representation implementation}
    In our experiment, we use a 2-level ellipsoid representation, which means that we construct the ellipsoid for the entire point cloud (Level 0) and each of its partition (Level 1). By default, we set $N_c = 36$, namely we decomposite the point cloud into $36$ partitions. For each ellipsoid latent space, we project the points into a $32\times32$ feature map ($M=32$). For point-wise features, we use the points 3D position in the ellipsoid coordinate frame; for ellipsoid-wise features, we use the orientation, the radii along each ellipsoid axle, and the position of ellipsoid center in the patent coordinate frame. 
    
    Taking advantage of the point cloud augmentation, we employ an end-to-end training scheme that rotates a point cloud, jitters, and generates an ellipsoid representation in each training step. For testing, we generate 1 representation for each object in all test sets.

\subsection{Network implementation}
    In the experiment, the default implementation of the EllipsoidNet modules is as follows: (1) the point-wise encoder is implemented as "$ResBlock(64)\rightarrow MaxPool(4\times4)\rightarrow ResBlock(128)\rightarrow MaxPool(4\times4)\rightarrow
    ResBlock(1024)\rightarrow MaxPool(2\times2)$";  (2) the ellipsoid-wise encoder is implemented as "$Conv(64)\rightarrow Conv(128)\rightarrow Conv(1024)$"; (3) the decoder module is implemented as "$Conv(512)\rightarrow Conv(256)$"; (4) the classification head is implemented as "$Fc(512)\rightarrow Fc(256) \rightarrow Fc(N_{CLS})$"; and (5) the segmentation head is implemented as "$Conv(128)\rightarrow Conv(N_{Seg})$". The $Conv$ denotes the 2D convolution layer, $Fc$ denotes the fully connected layer, $N_{CLS}$ denotes the number of object classes, and $N_{Seg}$ denotes the number of point semantic classes. A ReLU activation is added right after each ResBlock, convolution layer and fully-connected layers.
    
    The proposed representation method and neural network is implemented on a desktop equipped with an Intel Core i9-9820X CPU (3.3 Hz) and an NVidia GTX 1070 GPU. Key software packages include Python 3.7, Tensorflow 2.3 \cite{tensorflow2015-whitepaper}, Scipy 1.5.3 \cite{2020SciPy-NMeth} and Scikit-learn 0.23 \cite{scikit-learn}.
    
    The statistics of the implementation for ModelNet40 and ShapeNet are listed in Table \ref{tab:stistics}.
    
  \begin{table}[t]
		\centering
		\begin{tabular}{c|c|c}
			\toprule
			Target Dataset & ModelNet40 & ShapeNet \\
			\midrule
            Parameters (million) & 3.75 & 4.78 \\
            Representation time (ms) & 88.0 & 105.5 \\
            Network time (ms) & 8.4 & 13.0 \\
            Total time (ms) & 96.4 & 118.5 \\
			\bottomrule
		\end{tabular}
    	\caption{Statistics of method implementation.\label{tab:stistics}}
        \vspace{-3mm}
    \end{table}
    
\subsection{Training scheme}
    To train the network, we use the categorical cross entropy loss for both classification and segmentation tasks, and all  auxiliary losses are equally weighted. In addition, we employ the Adam \cite{kingma2014adam} optimizer with a $1\times 10^{-4}$ learning rate. We trained the network for 2000 epochs with a 32-sample batch for classification task and a 4-sample batch for segmentation task due to GPU memory limitation.

\subsection{Ablation study}
\bfsection{Impact of features}
    To study the impact of point-wise feature and ellipsoid-wise feature to the point cloud learning, we manage to train and evaluate the network using point-wise feature only, ellipsoid-wise feature only and a combination of both features. All three input options are evaluated in ModelNet40 and ShapeNet datasets. From Table \ref{tab:ablation_input} we observe that EllipsoidNet with either point-wise features or ellipsoid-wise features achieves better results than PointNet, a baseline method that parses the point cloud without any representation. In addition, EllipsoidNet with both feature inputs results in near state-of-the-art performance in both classification and segmentation tasks, which indicates that the method provides a robust point cloud representation for general point cloud learning.
    
    \begin{table}[t]
    		\centering
    		\begin{tabular}{c|c|c}
    			\toprule
    			Input Features & \begin{tabular}[c]{@{}c@{}} ModelNet40 \\ accuracy (\%)\end{tabular} & \begin{tabular}[c]{@{}c@{}} ShapeNet instance \\ mIoU (\%)\end{tabular} \\
    			\midrule
    			Ellipsoid-wise & 89.2 & - \\
                Point-wise & 91.3 & 83.4 \\
                
                Both & 92.7 & 86.9 \\
    			\bottomrule
    		\end{tabular}
    	\caption{Comparison of classification and segmentation result between Ellipsoid in different input features.}
    	\label{tab:ablation_input}
    	  \vspace{-3mm}
    \end{table}
    
\bfsection{Impact of decomposition}
    To study the impact of decomposition to the point cloud learning, we apply the ellipsoid representation with different number of partitions. In Table \ref{tab:point_usage_decomposition} we compare some metrics in the representation level, here we implement the neural network and explicitly compare the performance in classification and segmentation tasks. Table \ref{tab:ellipsoidnet_decomposition} shows that more point partitions result in better performance in both applications. The impact of decomposition to the point cloud applications shares the same trend as the impact to point cloud representation.

    \begin{table}[t]
    		\centering
    		\setlength\tabcolsep{3pt}
    		\begin{tabular}{c|c|c|c|c}
    			\toprule
    			Representation & Spherical & \multicolumn{3}{c}{Ellipsoid}\\
    			\midrule
    			\begin{tabular}[c]{@{}c@{}} Feature \\ map size\end{tabular}  & $1\times32^2$ & $16\times32^2$ & $25\times32^2$ & $36\times32^2$\\
    			\midrule
    			\begin{tabular}[c]{@{}c@{}} ModelNet40 \\ accuracy (\%)\end{tabular} & 73.9 & 87.3 & 89.2 & 92.5 \\
    			\midrule
    			\begin{tabular}[c]{@{}c@{}c@{}} ShapeNet \\ instance mIoU \\ (\%)\end{tabular} & 56.4 & 77.1 & 80.6 & 86.5 \\

    			\bottomrule
    		\end{tabular}
    	\caption{Comparison between ellipsoid representation methods with different decomposition sizes.}
    	\label{tab:ellipsoidnet_decomposition}
    \end{table}

\bfsection{Impact of latent space resolution}
    To study the impact of decomposition to the point cloud learning, we apply the ellipsoid representation with different feature map sizes. In Table \ref{tab:point_usage_imreso} we compare some metrics in the representation level, here we implement the neural network and explicitly compare the performance in classification and segmentation tasks. As shown in the table, $32\times 32$ feature map leads to the best result in both applications; a smaller feature map result in a significant drop in performance, perhaps it lost too much points during point cloud representation; a bigger feature map does not gain a significant improvement in point cloud learning, which shares the same trend of performance in point cloud representation in Table \ref{tab:ellipsoidnet_imreso}.
    
    \begin{table}[t]
    		\centering
    		\begin{tabular}{c|c|c|c}
    			\toprule
    			Representation & \multicolumn{3}{c}{Ellipsoid multi-scale}\\
    			\midrule
    			\begin{tabular}[c]{@{}c@{}} Feature \\ map size\end{tabular}  & $36\times16^2$ & $36\times32^2$ & $36\times64^2$\\
    			\midrule
    			\begin{tabular}[c]{@{}c@{}} ModelNet40 \\ accuracy (\%)\end{tabular} & 91.8 & 92.5 & 92.7 \\
    			\midrule
    			\begin{tabular}[c]{@{}c@{}} ShapeNet \\ instance mIoU (\%)\end{tabular} & 81.4 & 86.5 & 86.9 \\

    			\bottomrule
    		\end{tabular}
    	\caption{Comparison between ellipsoid representation methods with different feature map sizes.}
    	\label{tab:ellipsoidnet_imreso}
    	  \vspace{-3mm}
    \end{table}

\subsection{State-of-the-art performance comparison}

\bfsection{Quantitative comparison in ModelNet40}
    In Table \ref{tab:ModelNet} we compare the classification accuracy between ours and several existing works. Our work achieves near state-of-the-art accuracy in overall accuracy and the best in the mean accuracy for all shape classes. In addition, our work has much fewer floating point operations (FLOPs) than existing methods.
    
\begin{table}[t]
	\centering
    \begin{tabular}{c|c|c|c}
		    \toprule
            Method &   \begin{tabular}[c]{@{}c@{}}OA \\ (\%) \end{tabular} & \begin{tabular}[c]{@{}c@{}} mAcc \\ (\%) \end{tabular} & \begin{tabular}[c]{@{}c@{}}FLOPs \\ (million) \end{tabular}\\
            \midrule 
            PointNet \cite{qi2017pointnet} & 89.2 & 86.2 & 440\\
            PointNet++ \cite{qi2017pointnet++} & 90.7 & - & 1684 \\
            MVCNN \cite{su15mvcnn} & 90.1 & - & 62057 \\
            Pointwise-CNN \cite{hua2018pointwise} & 86.1 & 81.4 & -\\
            KPConv \cite{thomas2019kpconv} & 92.9 & - & 35.7 \\
            InterpCNN \cite{mao2019InterpCNN} & 93.0 & - & - \\
            RS-CNN \cite{liu2019rscnn} & \textbf{93.6} & - & 295\\
            Spherical CNNs \cite{esteves2018SphericalCNNs} & 88.9 & - & -\\
            DensePoint \cite{Liu_2019_DensePoint}  & 93.2 & - & 148\\
            $\Psi$-CNN \cite{li2018pointcnn} & 92.0 & 88.7 & 1581\\
            A-CNN \cite{komarichev2019acnn} & 92.6 & 90.3 & -\\
            KD-Net \cite{klokov2017kdnet} & 91.8 & 88.5 & -\\
            SO-Net \cite{li2018sonet} & 90.9 & 87.3 & -\\
            SCN \cite{xie2018scn} & 90.0 & 87.6 & - \\
            3DContextNet \cite{zeng20183dcontextnet} & 90.2 & - & - \\
            O-CNN \cite{wang2017ocnn}  & 89.9 & - & -\\
            ECC \cite{simonovsky2017ecc} & 87.4 & 83.2 & -\\
            
            \midrule 
            \textbf{Ours} & 92.7 & \textbf{91.4} & \textbf{7}\\
            \bottomrule
        \end{tabular}
	\caption{Result comparison  (\%) with recent works on ModelNet40. ‘OA’ represents the mean accuracy for all test instances and ‘mAcc’ represents the mean accuracy for all shape classes. The symbol ‘-’ means not given.}
	\label{tab:ModelNet}
	\vspace{-3mm}
\end{table}

\bfsection{Quantitative comparison in ShapeNet}
    In Table \ref{tab:Shapenet}, we compare the class mean intersection over union (class mIoU) and the instance mean intersection over union (instance mIoU) between ours and several existing 2D representation-based works. Our works achieves near state-of-the-art in both class mIoU and Instance mIoU. Again, our work takes much fewer floating point operations during network inference.
    
\begin{table}[ht]
		\centering
		\begin{tabular}{c|c|c|c}
		    \toprule
            Method & 
            \begin{tabular}[c]{@{}c@{}c@{}}class \\ mIoU \\ (\%)\end{tabular}
            &
            \begin{tabular}[c]{@{}c@{}c@{}}instance \\ mIoU \\ (\%)\end{tabular}
            &
            \begin{tabular}[c]{@{}c@{}}FLOPs \\ (million) \end{tabular}
            
            \\
            
            \midrule 
            PointNet \cite{qi2017pointnet}
            & 80.4 & 83.7 & 440\\
            
            Pointnet++ \cite{qi2017pointnet++}
            & 81.9 & 85.1 & 1684\\
            
            DGCNN \cite{wang2019dynamic}
            & 82.3 & 85.1 & 2768\\
            
            RS-CNN \cite{Liu_2019_RS_CNN}
            & 84.0 & 86.2 & 295\\
            
            DensePoint \cite{Liu_2019_DensePoint}
            & 84.2 & 86.4 & 651\\
 
            VV-Net \cite{Meng_2019_VV_Net}
            & 84.2 & \textbf{87.4} & -\\          
            
            PartNet \cite{Yu_2019_PartNet}
            & 84.1 & \textbf{87.4} & - \\       
            
            $\Psi$-CNN \cite{Lei_2019_Octree}
            & 83.4 & 86.8 & 1581\\

            PAN \cite{pan2019pointatrousnet}
            & 82.6 & 85.7 & -\\

            PointCNN \cite{li2018pointcnn}
            & 84.6 & 86.1 & 1682\\
            
            O-CNN \cite{wang2017ocnn}
            & \textbf{85.9} & - & -\\
            
            \midrule

            {\bf Ours}
            & 85.0 & 86.9 & \textbf{19}\\

			\bottomrule
		\end{tabular}
	\caption{ Result comparison  (\%) with recent works on ShapeNet. Numbers in red are the best in the column, and numbers in blue are the second best. class mIoU represents class mean intersection over union and instance mIoU represents instance mean intersection over union. The symbol ‘-’ means not given.}
	\label{tab:Shapenet}
\end{table}

\begin{figure}[t]
    \centering
    \includegraphics[width=0.9\columnwidth]{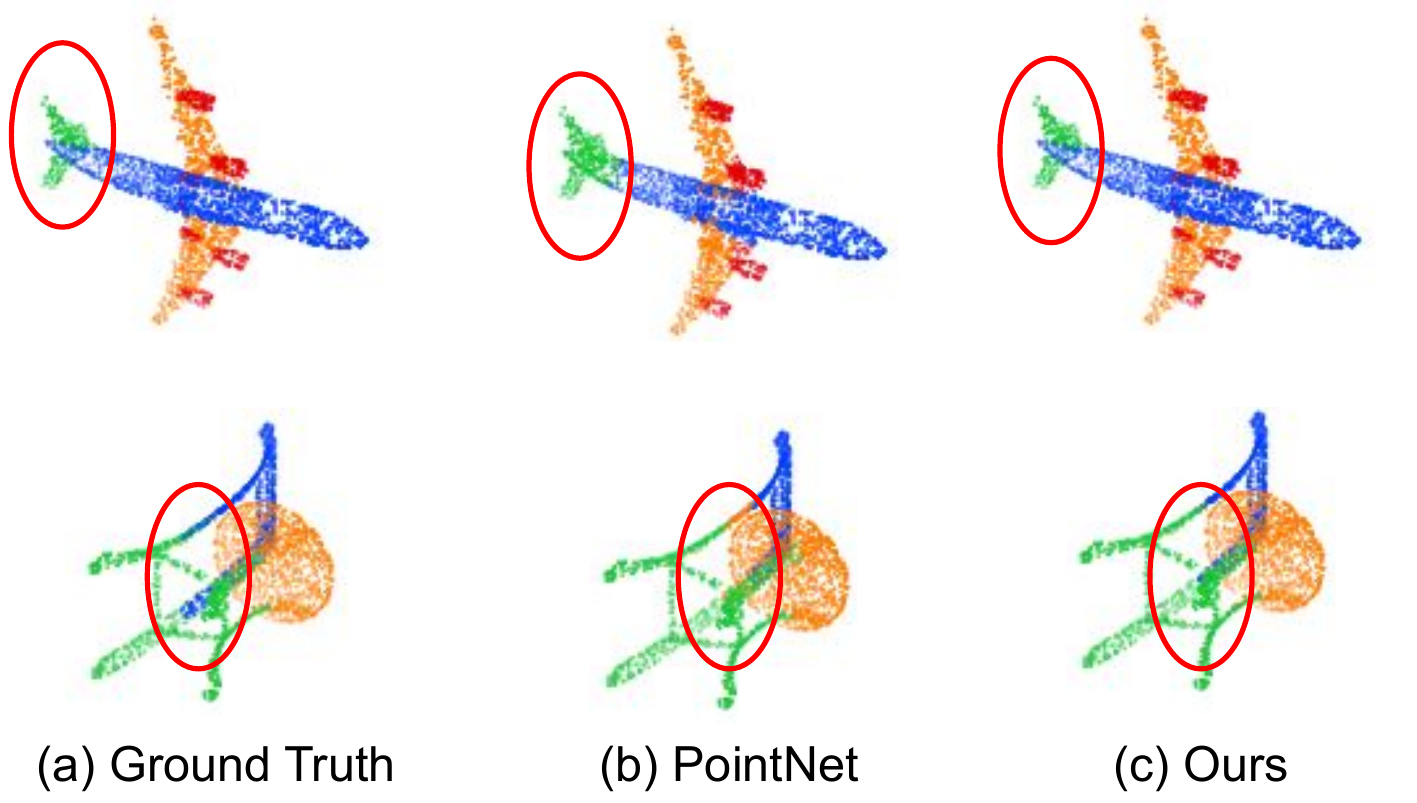}
    \caption{Qualitative comparison of point cloud segmentation between PointNet and ours. Points are colored by semantic segments. The different parts are emphasized using red circles. \label{fig:Qualitative_Comparison}}
    \end{figure}

\bfsection{Qualitative comparison}
    For qualitative comparison, we visualize several segmentation results in ShapeNet dataset together with the ground truth and results from PointNet, and ours in Figure \ref{fig:Qualitative_Comparison}. It is clearly seen that our method has better segmentation on small parts.
    
\section{Conclusion}
In this paper we address the problem of point cloud classification and segmentation via 2D representation learning. To this end, a novel point cloud representation method is proposed that hierarchically projects the point cloud to a ellipsoid space. The method effectively and efficiently projects the point cloud into 2D latent space for local feature extraction. In addition, a novel neural network architecture named EllipsoidNet is proposed to perform point cloud classification and segmentation from the proposed representations. We evaluate our method on ModelNet40 and ShapeNet, achieving near state-of-the-art performance on both data sets. In future works ellipsoid representation for point clouds scanned from the real scenes will be explored.

\section*{Acknowledgement}
Xinming Huang and Ziming Zhang were supported in part by NSF CCF-2006738.

\clearpage
{\small
\bibliographystyle{ieee_fullname}
\bibliography{egbib}
}

\end{document}